\pdfoutput=1
\documentclass[11pt]{article}

\usepackage[final]{acl}
\usepackage{times}
\usepackage{latexsym}
\usepackage{pifont}      % For symbols like \ding{}
\usepackage{amssymb}
\usepackage{float}

\usepackage{caption}
\usepackage{subcaption}

\usepackage[T1]{fontenc}
\usepackage[utf8]{inputenc}

\usepackage{microtype}

\usepackage{inconsolata}

\usepackage{graphicx}

\title{One Size Fits None: Rethinking Fairness in Medical AI}
\author{Roland Roller$^1$, Michael Hahn$^2$, Ajay Madhavan Ravichandran$^1$, Bilgin Osmanodja$^3$, \\ \textbf{Florian Oetke}$^4$, \textbf{Zeineb Sassi}$^5$, \textbf{Aljoscha Burchardt}$^1$, \textbf{Klaus Netter}$^5$, \textbf{Klemens Budde}$^3$, \\ \textbf{Anne Herrmann}$^{5,6}$, \textbf{Tobias Strapatsas}$^7$, \textbf{Peter Dabrock}$^2$, \textbf{Sebastian Möller}$^{1,8}$\\
  $^1$DFKI, $^2$Friedrich-Alexander-Universität Erlangen-Nürnberg,\\
  $^3$Charité - Universitätsmedizin Berlin, $^4$DNC Information Management GmbH, \\
  $^5$University of Regensburg,  $^6$University Hospital Regensburg,\\
  $^7$Asklepios Klinikum Harburg, $^8$TU Berlin}

\begin{document}
\maketitle
\begin{abstract}

%Machine learning (ML) models are increasingly used to support clinical decision-making. However, real-world medical datasets that are used to train these models are frequently noisy, incomplete, and imbalanced, resulting in performance disparities across different patient subgroups. This comparatively lower performance observed in certain subgroups may be examined through the lens of fairness, especially in cases where it exacerbates the disadvantages faced by already marginalized groups. In this work, we analyze several medical prediction tasks and show how model performance can vary depending on patient characteristics. While such ML models may provide reliable results for most of patients, we believe that reporting about performance on subgroup-level is essential before integrating them into clinical workflows. We highlight representative cases, discuss potential causes of these disparities, and outline how transparent reporting and awareness of training data limitations can help mitigate unintended harm. Our goal is to initiate a practical discussion around the subgroup-sensitive development and deployment of medical ML models and the interconnectedness of fairness and transparency.

Machine learning (ML) models are increasingly used to support clinical decision-making. However, real-world medical datasets are often noisy, incomplete, and imbalanced, leading to performance disparities across patient subgroups. These differences raise fairness concerns, particularly when they reinforce existing disadvantages for marginalized groups. In this work, we analyze several medical prediction tasks and demonstrate how model performance varies with patient characteristics.
%%%Michael: Vielleicht statt "perform well on average" etwas wie "show overall good performance". Weil sonst die Frage kommt, was denn "avarage" eigentlich ist --> außer das ist eben die Durchschnittsperson über alle Daten? Dann können wir es auch mit "average" lassen; nur philosophisch würde ich immer fragen, was denn die "Durchschnittsperson" sein soll.
While ML models may demonstrate good overall performance, we argue that subgroup-level evaluation is essential before integrating them into clinical workflows. By conducting a performance analysis at the subgroup level, differences can be clearly identified—allowing, on the one hand, for performance disparities to be considered in clinical practice, and on the other hand, for these insights to inform the responsible development of more effective models. 
%We highlight representative cases, discuss potential causes of disparities, and emphasize how transparent reporting and data awareness can help mitigate harm. 
Thereby, our work contributes to a practical discussion around the subgroup-sensitive development and deployment of medical ML models and the interconnectedness of fairness and transparency.

\end{abstract}

\section{Introduction}

Medical machine learning (ML) models are trained on datasets containing diverse patient characteristics. However, when certain subgroups are over- or underrepresented, models may show unequal performance, raising fairness concerns. Addressing such disparities requires evaluation across subgroups—ideally with an intersectional perspective that considers overlapping dimensions of disadvantage \cite{foulds2019intersectionaldefinitionfairness, wang2022intersectionality}.
This leads to the central question: \textbf{How should we address subgroup performance disparities in the context of fairness in medical ML?}

Fairness is a multifaceted concept that frequently arises in the context of machine learning systems. A common definition describes fairness in decision-making as the `absence of any prejudice or favoritism toward an individual or group based on their inherent or acquired characteristics' \cite{mehrabi2021survey}. Therefore, an ML system can be considered unfair if, despite the goal of achieving equally good performance across different subgroups, it exhibits substantial performance disparities. Those disparities often result from bias, for example through biased training data (data bias) or a biased algorithm itself (algorithmic bias). Both terms encompass various subtypes of bias, such as minority bias, missing data bias or cohort bias that can lead to a poorer performance for certain subgroups \cite{ueda2024fairness}.

%A core challenge is that real-world clinical data is often imbalanced and noisy—both in terms of outcome frequencies and subgroup representation—which can significantly impact model performance \cite{avati2021beds, elangovan2024principles}. Moreover, clinical evidence shows that patient outcomes and treatment responses differ systematically across populations, such as by sex or age. These differences reflect both biological variation and structural inequities in care access and treatment \cite{fradgley2015systematic, bruner2006reducing, patel2023equality}.

In machine learning, representation and performance disparities have been documented across modalities. For instance, large language models used in clinical settings may perpetuate stereotypes or marginalize certain identities when sociodemographic diversity is absent in training data \cite{alnegheimish2024can, lohse2024migration}. Similar issues arise in structured EHR modeling, where label noise and skewed sampling exacerbate subgroup-specific errors \cite{sivarajkumar2023fair, seyyed2020chexclusion}.

To address these challenges, prior work has taken different approaches. Some studies aim to improve dataset diversity or subgroup visibility in clinical training data \cite{rawat2024diversitymedqa, abraham2024improving}. Others propose fairness-aware optimization objectives or subgroup-specific tuning to reduce performance gaps \cite{sivarajkumar2023fair}. The importance of documentation and benchmarking has also been emphasized—especially in clinical imaging and foundation models—through standardized evaluation protocols across sensitive attributes \cite{jin2024fairmedfm}.

Our work contributes to this growing field by offering a structured analysis of subgroup variation across three real-world multimodal medical prediction tasks: mortality, triage, and graft failure, and advocating for routine reporting and subgroup validation as an integral part of the ethical assessment of medical ML model evaluation.

\section{Experiment}

We conduct our experiments on three multimodal clinical datasets, each containing textual data (e.g., clinical notes), structured static data (e.g., demographics), and, in two cases, time-series data (e.g., vital signs). All tasks involve patient-level predictions in distinct clinical settings.

%\cite{harutyunyan2019multitask}

\paragraph{Mortality} Based on the MIMIC-III \cite{johnson2016mimic} dataset from a US intensive care unit, this task involves predicting in-hospital mortality after the first 48 hours of admission \cite{yang2021leverage}. Data includes demographics, time-series vitals, and admission notes. It is framed as a binary classification and evaluated using AUC-ROC (ROC) and AUPRC (PRC).

\paragraph{Graft Failure} This dataset comes from a German transplant center and includes structured data (e.g., demographics, comorbidities), time-series labs and vitals, and clinical texts. The task is to predict graft failure within 360 days of each visit, using binary classification with ROC and AUPRC as metrics.

\paragraph{Triage} This dataset contains semi-structured ambulance records from a German emergency department, including structured features (e.g., vitals, pain score, Glasgow Coma Scale) and short text notes, describing the accident and situation of patient. The task is to classify patient urgency according to the Manchester Triage System (MTS), a multi-class classification problem evaluated using precision, recall, and F1 score.

\subsection{Methods}

We employ different machine learning models tailored to the characteristics of each dataset. The choice of method is influenced not only by the data modality and task complexity, but also by hardware constraints at the data hosting sites.

For \textbf{Mortality} prediction, we use a multimodal architecture that integrates irregular time-series and text data through interpolation-based embeddings and time-aware attention. Modalities are fused using interleaved self- and cross-attention layers, following the approach of \citet{zhang2022pm2f2n} and \citet{ravichandran2024xai}.
%For the \textbf{Mortality} prediction, we use a gated multimodal architecture combining a transformer for textual data, an LSTM for time-series inputs, and a feed-forward network for static structured features, following the approach described in \citet{yang2021leverage}. 
In the \textbf{Graft Failure} task, we apply a fast Gradient Boosting Regressor capable of handling static and time-series data as well as clinical notes, as described in \citet{roller2022evaluation}. For \textbf{Triage}, we apply a hybrid approach built around a transformer model for processing textual information, which is extended with a feed-forward network to integrate key structured features, as outlined in \citet{maschhur2024towards}. Additionally, expert rules are incorporated to better reflect aspects of the MTS and increase the recall for the most urgent classes. 

\subsection{Setup}

Each model is trained on a predefined training set and evaluated on a fixed test set, referred to as the \textit{reference test}. Using the same trained model, we then conduct a series of subgroup analyses by filtering the test set according to patient characteristics—for example, selecting only patients under 18 years old, or only female patients. Then, we compare the model’s performance on each subgroup against its performance on the full reference test set to investigate disparities across different patient groups.

%\subsection{Methods}
%
%In our experiments we rely on different machine learning models, which are capable to include the semi-structured data. However the selection of methods varies, which is partially do the data/project characteristics, but also hardware requirements in the location where a particular (sensitive) dataset is hosted. In case of \textbf{Mortality Prediction} using MIMIC data, we rely on a gated machine learning method combining a transformer (text), with an LSMT (time-variant) and a feed forward network (time-invariant), similarly as described in \textbf{X}. For the \textbf{Triage Classification} using the ambulance data, we rely hybrid machine learning models, building on a tranformer model, capable of integrating additionally most relevant structured information \textbf{by an additional feed forward network, as described in XXX}, combined with some expert rules. Finally, in case of \textbf{Graft Failure} using the nephrology data as apply a Gradient Boosting Regressor, integrating structured and unstructured data.
%
%\subsection{Setup}
%
%For each dataset we train our models on a given training dataset and evaluate it on a fixed test dataset, we refer to this set as reference test set. Using the exact same model, we conduct a subgroup analysis, filtering the test data according to different charactistics, e.g. targeting only patients below 18, or only on men. Next we compare the results on the subgroup in comparison to the reference test set and examine if we identify differences in performance.

\subsection{Subgroup Analysis Results}

Table~1-3 present results from our subgroup analysis across the three tasks. We observe that while overall performance is strong on the full test sets, notable variations emerge across subpopulations.
%Table~\ref{tab:MIMIC}, Table~\ref{tab:Nephrology} and Table~\ref{tab:KIBATIN_results} present preliminary results from our subgroup analysis across the three tasks. We observe that while overall model performance is strong on the full test sets, notable variations emerge across subpopulations.

\begin{table}[htbp!]
    \centering
    %\scriptsize % font size adjustment
        \centering
        \begin{tabular}{l|c}
        & Mortality \\\hline
        Test-Set & ROC - PRC \\ \hline
        \textbf{Reference} & 0.89 - 0.61  \\
        High Age (>75) & 0.86 - 0.59    \\ 
        Male & 0.90 - \colorbox{green!30}{0.65} \\ 
        Female & 0.88 - \colorbox{red!30}{0.57}  \\ 
        White & 0.89 - 0.62 \\ 
        Black & 0.86 - \colorbox{red!30}{0.45}  \\ 
        Asian & 0.91 - \colorbox{red!30}{0.56}  \\ 
        Hispanic & 0.97 - \colorbox{green!30}{0.77} \\ 
        Other & 0.90 - \colorbox{green!30}{0.70}  \\ 
        \hline
        \end{tabular}
        %\caption*{(a) Mortality}
        %\caption{Subgroup Analysis on Mortality Prediction.}
        
    \caption{Subgroup Analysis of the Mortality Task, using AUC-ROC (ROC) and Area under the Precision-Recall Curve (PRC).}
    \label{tab:MIMIC}
\end{table}

\begin{table}[htbp!]
    \centering
        \centering
        \begin{tabular}{l|c}
        & Graft Loss  \\\hline
        Test-Set & ROC - PRC  \\ \hline
        \textbf{Reference} & 0.94 - 0.55 \\
        Low Age & 0.96 - \colorbox{green!30}{0.72} \\
        High Age & 0.93 - 0.51  \\ 
        Male & 0.95 - \colorbox{green!30}{0.61} \\ 
        Female & 0.94 - \colorbox{red!30}{0.49} \\ 
        Donor Alive & \colorbox{green!30}{0.98} - \colorbox{green!30}{0.70} \\
        Donor Dead & 0.93 - 0.53 \\\hline
        \end{tabular}
        %\caption*{(b) Graft Failure}
        %\caption{Subgroup Analysis Graft Failure.}
        
    \caption{Subgroup Analysis of the Graft Failure Prediction Task, using AUC-ROC (ROC) and Area under the Precision-Recall Curve (PRC).}
    \label{tab:Nephrology}
\end{table}

%\begin{table}[htbp!]
%    \centering
%    \small
%    \begin{tabular}{l|cc}
%    & \multicolumn{2}{c}{Mortality}   \\\hline
%    Test-Set  & ROC & PRC \\ \hline
%    Reference Test & 0.89 & 0.61  \\
%    %Low Age &  & & &   \\
%    High Age & 0.89 & 0.60    \\ 
%    Male & 0.90 & \colorbox{green!30}{0.65} \\ 
%    Female & 0.88 & \colorbox{red!30}{0.57}  \\ 
%    White & 0.89 & 0.62 \\ 
%    Black & 0.86 & \colorbox{red!30}{0.45}  \\ 
%    Asian & 0.91 & \colorbox{red!30}{0.56}  \\ 
%    Hispanic & 0.97 & \colorbox{green!30}{0.77} \\ 
%    Other & 0.90 & \colorbox{green!30}{0.70}  \\ 
%    \hline
%    \end{tabular}
%    \caption{Subgroup Analysis on Mortality Prediction.}
%    \label{tab:MIMIC}
%\end{table}

\textbf{Mortality}: The model performs well overall (see Table \ref{tab:MIMIC}), but subgroup differences are notable in PRC, which are more sensitive to class imbalance. For instance, PRC is highest among male (0.65) and Hispanic patients (0.77), but substantially lower for women (0.57) and Black patients (0.45), suggesting a performance disparity, particularly in recall-sensitive settings. The score even further decreases for Black women to PRC=0.36 (not shown in the table).

%\begin{table}[htbp!]
%    \centering
%    \small
%    \begin{tabular}{l|cc}
%    &  \multicolumn{2}{c}{Graft Loss}  \\\hline
%    Test-Set  & ROC & PRC  \\ \hline
%    Reference Test & 0.94 & 0.55 \\
%    Low Age & 0.96 & \colorbox{green!30}{0.72} \\
%    High Age & 0.93 & 0.51  \\ 
%    Men & 0.95 & \colorbox{green!30}{0.61} \\ 
%    Women  & 0.94 & \colorbox{red!30}{0.49} \\ 
%    Donor Alive  & \colorbox{green!30}{0.98} & \colorbox{green!30}{0.70} \\
%    Donor Dead  & 0.93 & 0.53 \\\hline
%    \end{tabular}
%    \caption{Subgroup Analysis Graft Failure.}
%    \label{tab:Nephrology}
%\end{table}

\textbf{Graft Failure}: Similarly to above, subgroup differences are particularly notable in PRC (see Table \ref{tab:Nephrology}). Predictions are most reliable for younger patients (PRC=0.72), male patients (0.61), and recipients of organs from living donors (0.70). Performance drops for older patients, women, and cases with deceased donors—groups that may require additional calibration or targeted support.

\begin{table}[htbp!]
    \centering
    \tiny
    \begin{tabular}{l|ccc|ccc}
    & \multicolumn{3}{c|}{\textbf{Reference Test}}  & \multicolumn{3}{c}{Children (<18)}  \\\hline
    Labels  & Prec  & Rec & F1 & Prec & Rec & F1  \\ \hline
    %\multirow{3}{*}{E} & Affirmed & 0.88 & 0.97 & 0.93 & \textbf{0.97 }& \textbf{0.98} & \textbf{0.98}  \\
    Green & 0.53 & 0.40 & 0.46 & \colorbox{red!30}{0.47} & 0.42 & 0.44 \\
    Yellow & 0.63 & 0.47 & 0.54 & 0.65 & \colorbox{green!30}{0.56} & \colorbox{green!30}{0.60} \\
    Orange & 0.20 & 0.53 & 0.29 & \colorbox{green!30}{0.33} & \colorbox{red!30}{0.40} & \colorbox{green!30}{0.36} \\ 
    Red & 0.21 & 0.86 & 0.34 & \colorbox{green!30}{0.30} & \colorbox{red!30}{0.78} & \colorbox{green!30}{0.44} \\\hline\hline
    %Overall & \multicolumn{3}{c}{Micro-F1=; Macro-F1=}  & \multicolumn{3}{c}{Micro-F1=; Macro-F1=} \\\hline\hline
    & \multicolumn{3}{c|}{Male}  & \multicolumn{3}{c}{Female}  \\\hline
    Green & 0.53 & 0.39 & 0.45 & 0.53 & 0.42 & 0.47 \\
    Yellow & 0.63 & 0.48 & 0.55 & 0.63 & 0.46 & 0.53 \\
    Orange & 0.23 & \colorbox{green!30}{0.57} & 0.32 & 0.17 & 0.49 & 0.25 \\ 
    Red & \colorbox{green!30}{0.27} & 0.87 & \colorbox{green!30}{0.41} & \colorbox{red!30}{0.16} & 0.85 & \colorbox{red!30}{0.26} \\ \hline\hline
    %Overall & \multicolumn{3}{c}{Micro-F1=; Macro-F1=}  & \multicolumn{3}{c}{Micro-F1=; Macro-F1=} \\\hline\hline

    & \multicolumn{3}{c|}{High Age (>85)}  & \multicolumn{3}{c}{No Age}  \\\hline
    Green & \colorbox{green!30}{0.59} & 0.38 & 0.46 & 0.44 & \colorbox{red!30}{0.27} & \colorbox{red!30}{0.33} \\
    Yellow & 0.60 & \colorbox{green!30}{0.53} & 0.56 & \colorbox{red!30}{0.48} & \colorbox{red!30}{0.43} & \colorbox{red!30}{0.45} \\
    Orange & \colorbox{red!30}{0.13} & \colorbox{red!30}{0.44} & \colorbox{red!30}{0.20} & \colorbox{green!30}{0.45} & \colorbox{red!30}{0.45} & \colorbox{green!30}{0.45} \\ 
    Red & \colorbox{red!30}{0.16} & 0.88 & \colorbox{red!30}{0.27} & \colorbox{green!30}{0.36} & \colorbox{red!30}{0.67} & \colorbox{green!30}{0.47} \\ 
    %Overall & \multicolumn{3}{c}{Micro-F1=; Macro-F1=}  & \multicolumn{3}{c}{Micro-F1=; Macro-F1=} \\\hline\hline
    
%   & \multicolumn{3}{c|}{Middle Age}  & \multicolumn{3}{c}{No German}  \\\hline
%    Green &  &  &  &  &  &  \\
%    Yellow  &  &  &  &  &  &  \\
%    Orange  &  &  &  &  &  &  \\
%    Red  &  &  &  &  &  &  \\
%    Overall & \multicolumn{3}{c}{Micro-F1=; Macro-F1=}  & \multicolumn{3}{c}{Micro-F1=; Macro-F1=} \\
\hline
    
    \end{tabular}
    \caption{Subgroup Analysis on Triage Prediction}
    \label{tab:KIBATIN_results}
\end{table}

\textbf{Triage}: For children, less serious cases (red, orange) can be detected (lower recall). The overall performance (see Table \ref{tab:KIBATIN_results}) of male and female patients, instead, is roughly similar to the reference test set. Only the precision of the most serious class decreases for women, while it increases for men. In the case of old patients, above the model shows for red and orange a very strong performance drop. Finally, in cases where patient data does not include any age—and missing crucial information can occur frequently in real-world data of emergency care—we can see a drop in recall within all classes. Using solely the transformer-based machine learning model, we can see a similar pattern (see Appendix).

\section{Analysis}

\subsection{Medical Analysis} %\textbf{-> Tobias: \cite{van2008manchester}}
In the following, a brief analysis from a medical perspective is provided. 
\paragraph{Mortality} ICU settings offer rich data but cannot fully capture bedside clinical judgment, which is hard to textualize and prone to bias. Early ICU assessments, especially under stress, may introduce human biases that models can reproduce. Biological differences, such as higher baseline blood pressure in Black patients, may also skew mortality predictions if not properly accounted for.

\paragraph{Graft Loss} Graft loss risk is inversely linked to kidney function, estimated via creatinine-based eGFR. This is less reliable for frail patients with low muscle mass (common in elderly), possibly explaining reduced PRC. Gender bias may arise from the overrepresentation of men and the use of creatinine instead of sex-adjusted eGFR. Better performance in living-donor transplants may reflect generally improved outcomes, although this is harder to interpret due to many confounding variables.

\paragraph{Triage} Medically, triage is a challenging task, as the “correct” category often requires diagnostic confirmation, which is not considered for the given task. Even experienced nurses frequently mislabel cases, and paramedics may overtriage due to time pressure or to err on the side of caution. Known biases—such as overtriaging children and undertriaging cardiorespiratory symptoms—are reflected in model performance, which deviates most in children and the elderly. Overall, the label noise and potential misclassification limit the validity of model evaluation. Reliable ground truth is essential for meaningful ML applications in this context, but a manual analysis shows a large number of false triage labels in the real-world data (about 30\%).

\subsection{Technical Analysis}

\paragraph{Data Distribution}

All datasets are highly imbalanced with respect to the target events—such as mortality, graft failure, or red triage—which are rare and make machine learning tasks more challenging. Event frequency also varies across subgroups and between training and test sets, and subgroup sizes differ significantly, both in terms of total patients and percentage of target events. These factors can all impact model performance.

For instance, in the \textbf{Mortality} dataset, Asian patients make up only 2\% of the data (train and test), compared to 71\% for White patients, which may contribute to lower performance if subgroup-specific characteristics are important for prediction. However, despite representing 9\% of the population, the model performs worse on Black patients than on Asians (2\%) or Hispanics (3\%). Interestingly, the mortality rate for Black patients is only 9\%, compared to an overall average of 13\%. The gender ratio is roughly 55:45 (male:female), which could also contribute to performance differences.

Similar patterns are observed in the other two datasets (see Appendix), suggesting that subgroup composition likely affects model performance but cannot fully explain the observed disparities.

\paragraph{Significance} To examine concerns about spurious variation in small subgroups, where few positive cases can skew results, we conduct a one-sided nonparametric bootstrap hypothesis test on the \textbf{Mortality} task. We test if the model performed significantly better on one subgroup (A) than another (B). Overall, while we can see certain trends on particular subgroups of the \textbf{Mortality} data, the test found no significant performance differences between men and women, Hispanics and Whites, or Whites and Asians. However, the \textbf{model does perform significantly better for Whites compared to Blacks}\footnote{Corresponding confidence intervals as well as further details about the significance test, are reported in the Appendix.}.

%To assess whether the model performed significantly better on one subgroup (A) compared to another one (B), we applied a nonparametric one-sided bootstrap hypothesis test on the \textbf{Mortality} task. We computed the AUPRC for each subgroup across 1,000 bootstrap resamples of the test data (sampling with replacement), and then calculated the distribution of the pairwise difference ($AUPRCA_A$ – $AUPRCB_B$). A one-sided p-value was computed as the proportion of bootstrap differences less than or equal to zero. The difference was considered statistically significant if the p-value was below 0.05.

%This test also helps address concerns about spurious performance variation in subgroups with limited sample sizes. For instance, if a subgroup contains only a few positive cases that happen to be misclassified, performance can appear artificially low. By bootstrapping, we estimate the variability of subgroup performance due to sampling, and test whether observed differences are likely to reflect true model bias or merely random chance.

%Although Table 1(a) shows clear tendencies, based on the hypothesis test we cannot conclude that the model works significantly better on men, compared to women, or with Hispanics, compared to Whites, and neither Whites compared to Asians. What we could find out, however, is that the model works in fact significantly better for Whites, compared to Blacks. Confidence intervals can be found in Table \ref{tab:MIMIC_cinf_interval} in the Appendix.

\section{Discussion}

%\colorbox{red!30}{1. bias is there, 2. sometimes also good, as it starts new medical research? 3. Suggest information leaflet}

Our results highlight the variability of ML model performance across patient subgroups on different multimodal datasets in multiple tasks. While overall metrics may suggest good performance, a closer look reveals that \textbf{models can underperform for specific subgroups}, such as older patients, individuals from certain ethnic groups, but also patients with lower data quality or a particular transplant. This poses a potential risk, particularly in clinical decision-making, where complex and difficult decisions must be made for vulnerable patient populations. 

%These disparities are not surprising, given the typical characteristics of medical datasets: they are often incomplete, imbalanced, and reflective of systemic inequalities. For instance, smaller subgroups may be underrepresented during training, leading to lower predictive quality. There is considerable evidence to suggest that many patient subgroups, including women and patients with a migration background, are underrepresented in clinical research\footnote{NIHR (2020) Improving Inclusion of under-Served Groups in Clinical Research: Guidance from the NIHR-INCLUDE Project. Accessed March 7, 2025. www.nihr.ac.uk/documents/improving-inclusion-of-under-served-groups-in-clinical-research-guidance-from-include-project/25435} \cite{patel2023equality} and only have limited access to optimal care \cite{fradgley2015systematic,bruner2006reducing}. Also, it has been shown that LLM-generated clinical vignettes often fail to represent sociodemographic diversity potentially reproducing racial and gender stereotypes and unconscious biases \cite{lohse2024migration}. Moreover, the complexity of medical decision-making may be insufficiently captured by available features, further amplifying bias when models generalize based on limited patterns.

As we have shown, fairness can be understood as the requirement that different subgroups should exhibit similar performance and that the model should not `favor' any particular subgroup. However, in order to be fair and to pursue the goal of achieving equal performance across all subgroups, transparency is essential. First, it must be recognized that the model performs differently across different subgroups. With this knowledge of the subgroup-specific performance disparities a particular model can still be used—especially since, in many real-world scenarios, achieving fairness in the sense of identical performance for all subgroups may not be feasible. But for that to be responsible, it is important that these \textbf{models are accompanied by documentation} similar to an `\textit{information leaflet}' or a `\textit{package insert}' \cite{samhammer2023klinischeentscheidungsfindung, ott2022transparenthuman} that includes subgroup-level performance metrics, an overview of the training data distribution, and disclaimers when certain subgroups are likely underrepresented. The EU AI Act even demands a respective documentation for high-risk AI systems \cite{euai2024article13}. To this end, best practices and standards for reporting subgroup performance need to be developed. Such information can then guide clinicians in interpreting predictions, managing uncertainty, and identifying when to override or ignore model outputs.

At the same time, this \textbf{transparency must not become a substitute for fairness}, allowing largely unfair and biased models to be used uncritically and thereby reinforcing existing inequalities. Rather, transparency and fairness must be closely intertwined, with the recognition of poorer performance for certain subgroups prompting targeted efforts to improve outcomes specifically for those groups. 

%Our findings call for future research into strategies for mitigating these disparities. Potential approaches include subgroup-specific calibration, data augmentation, fairness-aware training objectives, and post-hoc correction mechanisms. 

Ultimately, the goal should not be to prevent the use of models that do not perform equally for all possible subgroups, but to ensure they are used with awareness, and that this insight is used to improve the model specifically for those disadvantaged groups. A \textbf{biased model with clear warnings and transparent evaluation may still bring benefit in clinical practice}, especially in settings where no decision support exists otherwise. However, it is precisely this transparency enabled by subgroup analysis that can help further improve the model or even develop a new model specifically for those subgroups that are otherwise underrepresented. Finally, the knowledge about surprising performance discrepancies across patient subgroups can also \textbf{trigger further research, as the underlying causes could also be medical rather than solely data-driven}.  %Transparency must not become a justification for deploying unfair models, and the responsible use of ML models does not end with transparently documenting performance disparities across different subgroups.

\section{Conclusion}

In this paper, we presented a pragmatic perspective on fairness challenges in medical machine learning. Through empirical subgroup analyses on three diverse clinical tasks, we showed that performance disparities across patient populations are not only common but often hidden by aggregate metrics. Since `one size fits all' solutions, where ML models aim but fail to perform equally across all subgroups, are rarely adequate in real-world scenarios, we have demonstrated the importance of linking fairness and transparency: making biases visible, reporting subgroup-specific performance, and acknowledging data limitations. Also, we need further efforts to help overcome access barriers to clinical research and optimal care, as this would also help to improve medical datasets used to develop and train fair models. Likewise, best practices and standards for evaluating and reporting subgroup performance need to be developed. This transparency serves two purposes: it allows physicians to weigh in on the model’s performance across subgroups for clinical decision-making, and at the same time, it enables targeted optimization of the model for those groups that are currently disadvantaged. In doing so, we can foster more responsible use of ML models in healthcare.

%Future work should explore methods to reduce subgroup disparities and evaluate their practical impact. Ultimately, fairness in medical AI must be seen as a process, not a checkbox — one that combines technical effort with continuous reflection, reporting, and engagement with clinicians and patients alike.

%\newpage
\section*{Bias Statement}
We define the considered biases as performance disparities across patient subgroups based on particular characteristics, such as age, gender, ethnicity, but also data quality or donor. These biases are harmful because they can lead to misdiagnosis or suboptimal care for marginalized groups—for example, by underpredicting mortality risk in older or female patients, or by providing less accurate triage classifications for children. Such disparities may reinforce existing inequalities in clinical care.

Our work demonstrates that these behaviors arise due to underrepresentation in training data, label noise, and missing information in real-world medical datasets. We advocate for transparent subgroup reporting, which enables clinicians and developers to identify when model outputs should be questioned or overridden. In doing so, we aim to promote safer, more equitable AI integration into clinical practice.

\section*{Limitations}

Our subgroup analyses are exploratory and based on straightforward demographic or clinical splits (e.g., age, gender), without a principled approach to subgroup formation. 
%%%Michael: Ich bin unsicher, ob wir sagen wollen, dass wir intersektionale Effekte nicht berücksichtigen, weil wir das ja bspw. mit dem Beispieln von Schwarzen Frauen tun. Vielleicht könnten wir das hier einfach weglassen und positiv für die zukünftige Forschung hervorheben.
%This may overlook relevant intersectional effects or disadvantage finer-grained subpopulations. 
Future work should explore systematic strategies for identifying meaningful subgroups, particularly to ensure fair model performance across underrepresented or multiply marginalized patient groups by applying a decidedly intersectional perspective. Additionally, while we account for performance differences, we do not explicitly quantify uncertainty or statistical significance across all datasets and subgroups. The clinical datasets we rely on also exhibit label noise, missing values, and potential bias in documentation practices (e.g., in triage labels or notes), which can affect both model training and evaluation. Finally, generalizability may be limited, as two datasets are from Germany and one from a single US hospital.

\section*{Acknowledgments}
%Bei PRIMA-AI habe ich das "C" am Ende weggenommen, weil ja das Gesamtprojekt beteiligt ist (geht das?) und noch EmpkinS in Erlangen ergänzt.
The project has received funding from the Federal Ministry of Research, Technology and Space through the projects KIBATIN (16SV9040) and PRIMA-AI (01GP2202), from the Deutsche Forschungsgemeinschaft (DFG, German Research Foundation) — SFB 1483 — Project-ID 442419336, EmpkinS and the Federal Joint Committee of Germany (Gemeinsamer Bundesausschuss) as
part of the project smartNTX (01NVF21116).

\bibliography{custom}

%\newpage
%\newpage
\appendix

\section{Appendix}
\label{sec:appendix}

\subsection{Triage Prediction using ML Model}
Table \ref{tab:KIBATIN_DFKI} represents the results on the \textbf{Triage} dataset using only the transformer-based machine learning model - opposed to the model in Table \ref{tab:KIBATIN_results}, which optimizes on recall, and integrates expert knowledge.

\begin{table}[H] %htbp!]
    \centering
    \tiny
    \begin{tabular}{l|ccc|ccc}
    & \multicolumn{3}{c|}{Reference Test}  & \multicolumn{3}{c}{Children (<18)}  \\\hline
    Labels  & Prec  & Rec & F1 & Prec & Rec & F1  \\ \hline
    %\multirow{3}{*}{E} & Affirmed & 0.88 & 0.97 & 0.93 & \textbf{0.97 }& \textbf{0.98} & \textbf{0.98}  \\
    Green & 0.52 & 0.28 & 0.37 & \colorbox{red!30}{0.44} & 0.29 & 0.35 \\
    Yellow & 0.58 & 0.64 & 0.61 & 0.61 & \colorbox{green!30}{0.69} & 0.64 \\
    Orange & 0.22 & 0.48 & 0.30 & \colorbox{green!30}{0.27} & \colorbox{red!30}{0.36} & 0.31 \\ 
    Red & 0.44 & 0.45 & 0.45 & \colorbox{green!30}{0.50} & \colorbox{red!30}{0.28} & 0.36 \\ \hline\hline
    & \multicolumn{3}{c|}{Male}  & \multicolumn{3}{c}{Female}  \\\hline
    Green & 0.54 & 0.27 & 0.36 & 0.51 & 0.29 & 0.37 \\
    Yellow & 0.58 & 0.65 & 0.61 & 0.59 & 0.64 & 0.61 \\
    Orange & 0.23 & 0.50 & 0.32 & 0.21 & 0.46 & 0.29 \\ 
    Red & \colorbox{green!30}{0.49} & 0.43 & 0.46 & \colorbox{red!30}{0.38} & 0.47 & 0.42 \\ \hline \hline

    & \multicolumn{3}{c|}{High Age (>85)}  & \multicolumn{3}{c}{No Age}  \\\hline
    Green & \colorbox{green!30}{0.60} & 0.25 & 0.35 & \colorbox{red!30}{0.46} & \colorbox{red!30}{0.20} & \colorbox{red!30}{0.28} \\
    Yellow & 0.56 & \colorbox{green!30}{0.70} & 0.62 & \colorbox{red!30}{0.50} & \colorbox{green!30}{0.75} & 0.60 \\
    Orange & 0.18 & 0.46 & 0.26 & \colorbox{red!30}{0.11} & \colorbox{red!30}{0.09} & \colorbox{red!30}{0.10} \\ 
    Red & \colorbox{red!30}{0.32} & 0.44 & \colorbox{red!30}{0.37} & \colorbox{green!30}{0.67} & \colorbox{red!30}{0.33} & 0.44 \\ \hline
    \end{tabular}
    \caption{Subgroup Analysis on Triage Prediction with ML model}
    \label{tab:KIBATIN_DFKI}
\end{table}

\begin{table*}[htbp!]
    \centering
    \footnotesize
    \begin{tabular}{l|cccc}
      & \multicolumn{2}{c}{Size Train} & \multicolumn{2}{c}{Size Test} \\ 
    Subgroups  & Freq. Absolute & Percent & Freq. Absolute & Percent \\ \hline
    Reference Test & 14068 (1852) & 100\% (13\%) & 3099 (359) & 100\% (12\%)\\
    %Low Age &  & & &   \\
    
    High Age (>75) & 3776 (664) & 27\% (17\%) & 834 (24) &  27\% (3\%)  \\ 
    %High Age & 8219 (1226) & 1768 (219) & &   \\ 
    Male & 7794 (997) & 55\% (13\%) & 1732 (193) & 56\% (11\%) \\ 
    Female & 6274 (855) & 45\% (14\%) & 1367 (166) & 44\% (12\%) \\ 
    White & 10002 (1276) & 71\% (13\%) & 2229 (253) & 72\% (11\%) \\ 
    Black & 1285 (112) & 9\% (9\%) & 270 (24) & 9\% (9\%) \\ 
    Asian & 335 (45) & 2\% (13\%) & 61 (9) & 2\% (15\%) \\ 
    Hispanic & 451 (36) & 3\% (8\%) & 106 (8) & 3\% (8\%) \\ 
    Other & 1995 (383) & 14\% (19\%) & 433 (66) & 14\% (15\%) \\ 
    \hline
    \end{tabular}
    \caption{Frequency of patients of Mortality task in subgroups within train and test. }
    \label{tab:MIMIC_dist}
\end{table*}

\begin{table*}[htbp!]
    \centering
    \small
    \begin{tabular}{l|cccccc}

    Labels  & All  & Children & Male & Female & High Age & No Age   \\ \hline
    Green & 3134 (34.82\%) & 293 (30.58\%) & 1492 (34.31\%) & 1638 (35.35\%) & 700 (38.76\%) & 30 (32.97\%) \\
    Yellow & 4951 (55.00\%) & 518 (54.07\%) & 2366 (54.42\%) & 2572 (55.50\%) & 977 (54.10\%) & 44 (48.35\%) \\
    Orange & 792 (8.80\%) & 129 (13.47\%) & 413 (9.50\%) & 378 (8.16\%) & 113 (6.26\%) & 11 (12.09\%) \\
    Red & 124 (1.38\%) & 18 (1.88\%) & 77 (1.77\%) & 46 (0.99\%) & 16 (0.89\%) & 6 (6.59\%) \\ \hline\hline
    percent & (9001) 100\%  & 10.64\% & 48.31\% & 51.48\% & 20.02\%& 1.01\%\\
    \end{tabular}
    \caption{Data Distribution Triage Prediction, showing the distributions of the four labels \textit{green}, \textit{yellow}, \textit{orange} and \textit{red} across the subgroups, as well as the overall percentage of patients of that group in the overall dataset. }
    \label{tab:KIBATIN_distribution}
\end{table*}

\begin{table*}[htbp!]
    \centering
    \small
    \begin{tabular}{l|cccc}
      & \multicolumn{2}{c}{Train}  & \multicolumn{2}{c}{Test}   \\ \hline
    Subgroups & Patients & Data Points (Target) & Patients & Data Points (Target) \\
    Reference Test & 1552 & 10321 (727) & 297 & 43945 (2813) \\
    Low Age (<30) & - & 1025 (65)  & - & 4335 (322) \\
    High Age (>75) & - & 449 (94) & - & 1401 (120)   \\ 
    Male  & 953 & 6391 (404) & 183 & 27425 (1690) \\ 
    Female  & 599 & 3930 (323) & 114 & 16520 (1123) \\ 
    Donor Alive  & 533 & 3427 (170) & 97 & 13085 (703) \\
    Donor Dead  & 1019 & 6894 (557) & 200 & 30860 (2110) \\\hline
    \end{tabular}
    \caption{Graft Failure: Frequency of patients and datapoints in train in test set within one split of cross validation}
    \label{tab:Nephro}
\end{table*}

\begin{table*}[htbp!]
    \centering
    \small
    \begin{tabular}{l|cc}
    Subgroups  & Mean & Confidence Interval \\ \hline
    Middle Age (>45) & 0.6802 & [0.5639, 0.7830] \\ 
    High Age (>75) & 0.5957 & [0.5050, 0.6830] \\ 
    Male & 0.6554 & [0.5920, 0.7170] \\ 
    Female & 0.5801 & [0.5039, 0.6610] \\ 
    White & 0.6183 & [0.5600, 0.6730] \\ 
    Black & 0.4444 & [0.2320, 0.6341] \\ 
    Asian & 0.5891 & [0.2608, 0.9351] \\ 
    Hispanic & 0.7642 & [0.4290, 0.9851] \\ 
    Other & 0.6976 & [0.5830, 0.7991] \\ 
    \hline
    \end{tabular}
    \caption{Mortality Prediction: Confidence intervals (95\%) of AUPRC based on 1,000 iterations of a one-sided bootstrap hypothesis test.}
    \label{tab:MIMIC_cinf_interval}
\end{table*}

\subsection{Data Point and Patient Frequencies}

Due to limited space and due to the fact that the main text can be easily understood without the detailed tables about data points and patient frequencies, we present them here in the Appendix (Tables \ref{tab:MIMIC_dist}, Table \ref{tab:Nephro} and \ref{tab:KIBATIN_distribution}).

Table \ref{tab:MIMIC_dist} presents the distribution of patients across subgroups for the mortality prediction task in the training and test sets. The table shows the absolute number of patients per subgroup, with the number of deaths in parentheses. Additionally, it reports the percentage of patients in each subgroup relative to the total dataset, and the mortality rate within each subgroup (i.e., percentage of deaths among subgroup members, also shown in parentheses).

Table \ref{tab:Nephro} shows the distribution of patients and their datapoints over time within training and test data of one split. The original split into training and test for the cross validation did not take possible subgroup information into account. Instead the split for the cross validations was conducted based on an equal distribution of patients with their number of included data points. Note, as kidney disease is a life long treatment, and our electronic patient record contains data over a long time, we make a forecast each time we insert new data for a patient (e.g. regular checkup or hospitalization).

Table \ref{tab:KIBATIN_distribution} presents the label distribution in the \textbf{Triage} dataset. Each column represents a subgroup, showing its proportion within the overall dataset (\textit{percent}) and the number of patient cases per triage class within that subgroup, along with the corresponding percentages relative to the subgroup total.

\subsection{Significance Test on Mortality}
To test if the model performed significantly better on one subgroup (A) than another (B) in the \textbf{Mortality} task, we ran a one-sided nonparametric bootstrap hypothesis test. We computed PRC for each subgroup across 1,000 bootstrap resamples (sampling with replacement) and calculated the distribution of the pairwise difference ($PRC_A$ – $PRC_B$). A one-sided p-value was then derived as the proportion of differences 	$\leq$ 0. Differences were considered significant at p~$<$~0.05.

This method also mitigates concerns about spurious variation in small subgroups, where few positive cases can skew results. Bootstrapping estimates performance variability due to sampling and helps distinguish real model bias from chance.

In this context, Table \ref{tab:MIMIC_cinf_interval} presents the confidence intervals of the different subgroups of the \textbf{Mortality} dataset. In many cases, particularly for the smaller subgroups, the confidence intervals show a large performance fluctuations.

%\begin{table*}[htbp!]
%    \centering
%    \tiny
%    \begin{tabular}{l|ccccccc}
%
%    Labels  & All  & Children & Male & Female & High Age & No Age & No German   \\ \hline
%    Green & 3233 (33\%) & \colorbox{green!30}{302 (28\%)} & 1549 (33\%) & 1679  (34\%) & & & \\
 %   Yellow & 5345 (55\%) & 577 (53\%) & 2537 (54\%) & 2794  (56\%)  & & & \\
%    Orange & 916 (9\%) & \colorbox{red!30}{168 (16\%)} & 487 (10\%) & \colorbox{green!30}{426 (9\%)}  & & & \\ 
%    Red & 211 (2\%)& \colorbox{red!30}{32 (3\%)} & \colorbox{red!30}{129 (3\%)} & \colorbox{green!30}{81 (2\%)}  & & & \\ \hline\hline
%    percent & 100\% & 11.12\% & 48.45\% & 51.31\%  & & & \\
%    \end{tabular}
%    \caption{OLD: Data Distribution Triage Prediction, percentage increase/decrease from the base (All) by more than 10\% are indicated in green/red}
%    \label{tab:KIBATIN_distribution}
%\end{table*}

\end{document}